\documentclass[conference]{IEEEtran}
\IEEEoverridecommandlockouts

\usepackage{cite}
\usepackage{amsmath,amssymb,amsfonts}
\usepackage{algorithmic}
\usepackage{graphicx}
\usepackage{textcomp}
\usepackage{xcolor}
\def\BibTeX{{\rm B\kern-.05em{\sc i\kern-.025em b}\kern-.08em
    T\kern-.1667em\lower.7ex\hbox{E}\kern-.125emX}}

\usepackage{easyeqn}

\def\url#1{\expandafter\string\csname #1\endcsname}

\def\RR{\mathbb{R}}

\def\ev{\mathbf{e}}
\def\av{\mathbf{a}}
\def\t{\mathrm{\times}}
\def\T{\mathrm{T}}

\def\EC{\mathcal{E}}

\def\LC{\mathcal{L}}
\def\DC{\mathcal{D}}

\def\EC{\mathcal{E}}
\def\VC{\mathcal{V}}
\def\GC{\mathcal{G}}

\def\bv{\textbf{b}}

\def\nsize{n}
\def\encoder{\EC}
\def\decoder{\DC}
\def\adjacencyMatrix{\mathbf{A}}
\def\nodeSet{\VC}
\def\edgeSet{E}

\def\Loss{\LC}
\def\similarity{s}
\def\similarityVector{\mathbf{s}}
\def\similarityMatrix{\mathbf{S}}
\def\embeddingMatrix{\mathbf{W}}
\def\histPos{P^+}
\def\histNeg{P^-}
\def\encoderParams{\Theta}

\begin{document}

\title{Constructing Graph Node Embeddings via Discrimination of Similarity Distributions
\thanks{The research was supported by the Russian Science Foundation grant (project 14-50-00150).}
}

\author{\IEEEauthorblockN{Stanislav Tsepa}
\IEEEauthorblockA{\textit{Skolkovo Institute of Science and Technology} \\
(Skoltech) \\
\textit{Moscow Institute of Physics and Technology} \\
\textit{Institute for Information Transmission Problems of RAS} \\
Moscow, Russia \\
s.tsepa@skoltech.ru}
\and
\IEEEauthorblockN{Maxim Panov}
\IEEEauthorblockA{\textit{Skolkovo Institute of Science and Technology} \\
(Skoltech) \\
\textit{Institute for Information Transmission Problems of RAS} \\
Moscow, Russia \\
m.panov@skoltech.ru}
}

\maketitle

\begin{abstract}
  The problem of unsupervised learning node embeddings in graphs is one of the important directions in modern network science. In this work we propose a novel framework, which is aimed to find embeddings by \textit{discriminating distributions of similarities (DDoS)} between nodes in the graph. The general idea is implemented by maximizing the \textit{earth mover distance} between distributions of decoded similarities of similar and dissimilar nodes. The resulting algorithm generates embeddings which give a state-of-the-art performance in the problem of link prediction in real-world graphs.
\end{abstract}


\section{Introduction}
  The quality of machine learning methods largely depends on the particular representation (or features) chosen for the data. The majority of modern machine learning methods work with the objects represented as numerical vectors. In some problems, such as object recognition on images and video, speech recognition and natural language processing, the initial feature space formally can be considered as a vector space. The obstacle is that it has a complex structure and a very high dimension which requires the construction of methods to transform the original representation into a more concise and informative one. Problems with objects of a discrete nature (in particular, with graphs) also require that informative continuous numerical representations have to be found. Recently network representation learning attracted a lot of attention which lead the development of many new methods (e.g., see recent reviews~\cite{Hamilton2017,Cai2018}).

  The essence of the network representation learning  problem (or embedding problem) is to represent a graph, a subgraph or a node as a point in low-dimensional Euclidean space; in such a form they can be further used in traditional machine learning pipelines. In what follows we will focus on node embeddings. A usual assumption is that nodes can be represented in a space with dimension \(d \ll \nsize\), where \(\nsize\) is the number of nodes in the graph. The obtained embeddings can be further used for node classification~\cite{Grover2016}, community detection~\cite{Fortunato2010}, link prediction~\cite{Backstrom2011} or visualization~\cite{Oliveira2017}.

  The majority of now existing graph embedding algorithms focus on proximity preserving embeddings, in which the nodes positioned in a close network proximity are considered to be similar and their embedding points should be placed close to each other in embedding space. We base our research on this assumption and, meanwhile, use the area of representation learning for images as source of inspiration. Recently the histogram loss approach was proposed~\cite{hist_loss}, where the embeddings are learned by minimizing a certain distance between inter- and intra-class similarity distributions. We extend their approach to graphs by introducing a more suitable distance between the distributions which is inspired by a Wasserstein distance. We demonstrate the efficiency of the proposed approach on a series of experiments with real-world graphs.
    
\section{Related work}
  Currently the methods of learning mode representations in graphs are rapidly developing. The earlier approaches are based on classical dimension reduction and on finding node embeddings via matrix factorization~\cite{Belkin2002,HOPE}. Matrix factorization can be really time consuming for large graphs, and a usual way to speed-up embedding learning is to use random walks over the graph. The idea is to find such vector representations of vertices which can describe well the probabilities of particular vertex sequences in these walks. This idea is the base for the whole family of methods, including DeepWalk~\cite{deepwalk}, LINE~\cite{Tang2015} and node2vec~\cite{Grover2016}. Note that all these methods use internally the algorithm word2vec~\cite{Mikolov2013}, which is based on the optimization of the logistic-like loss and was initially proposed for word embeddings. Further developments of random walk-based methods mainly focus on various schemes of random walk through the graph, which allow them to take into account various structural features of graphs. Recently, many attempts were made to adopt neural networks to graph-structured data (see~\cite{sdne,Cao2016} among many others). In the next section we specify the considered framework and further discuss some related embedding learning approaches.

\section{Learning graph embeddings in unsupervised way}
  We start by formalizing the problem of node representation learning in graphs following~\cite{Hamilton2017}. We observe an undirected and unweighted graph \(\GC = (\nodeSet, \edgeSet)\) with \(\nsize = |V|\) vertices and denote by \(\adjacencyMatrix \in \{0, 1\}^{\nsize \t \nsize}\) the adjacency matrix of graph \(\GC\) with \(\nsize\) nodes. We also assume that certain similarity matrix is given \(\similarityMatrix = \bigl[\similarity_{ij}\bigr]_{i, j = 1}^{\nsize}\), where the value \(\similarity_{ij}\) determines how similar the nodes \(i\) and \(j\) are, and, respectively, how close their embeddings should be. The considered framework consists of several important parts:
  \begin{enumerate}
    \item encoder function which maps nodes to the latent representations
      \begin{EQA}[c]
        \encoder \colon \nodeSet \to \RR^{d};
      \end{EQA}

    \item decoder function which maps pairs of node embeddings to node proximity measure
      \begin{EQA}[c]
        \decoder \colon \RR^{d} \t \RR^{d} \to \RR;
      \end{EQA}

    \item loss function \(\Loss\) which measures how close reconstructed proximity values \(\decoder(\encoder_i, \encoder_j)\) are to the corresponding reference values \(\similarity_{ij}\).
  \end{enumerate}

  In what follows we discuss different types of encoders and loss functions as the choice of them largely distinguishes modern embedding learning algorithms.

\subsection{Loss functions}
  We describe two important types of losses considered in the literature.

\subsubsection{Pairwise loss}
  in this case, the goal of optimization is to minimize the sum of reconstruction errors for pairwise similarities of nodes \(\ell\bigl(\decoder(\encoder_i, \encoder_j), \similarity_{ij}\bigr)\):  
  \begin{EQA}[c]
    \LC(\encoderParams) = \sum_{(i, j) \in \Omega} \ell\bigl(\decoder(\encoder_i, \encoder_j), \similarity_{ij}\bigr),
  \end{EQA}
  where decoder is usually considered to be non-parametrized and optimization is done over the parameters of encoder \(\encoderParams\).

\subsubsection{Nodewise autoencoder loss}
  this approach was recently proposed by~\cite{Tsitsulin2018}, where the authors consider the similarity of node embeddings as distributions:
  \begin{EQA}[c]
    \LC(\encoderParams) = \sum_{i = 1}^{\nsize} \ell\bigl(\decoder(\encoder_i, \cdot), \similarityVector_i\bigr),
  \end{EQA}
  where \(\decoder(\encoder_i, \cdot)\) and \(\similarityVector_i\) are vectors of decoded similarities and observed similarities of \(i\)-th node respectively. One of the possible choices for the loss function is Kullback-Leibler divergence. A similar approach is considered in~\cite{Wang2016,Cao2016}, where autoencoders were constructed considering the classical \(L_2\) loss.

\subsection{Encoders}
  The standard approach is the so-called embedding lookup, where
  \begin{EQA}[c]
  \label{direct_embedding}
    \encoder_i = \mathbf{Z} \ev_i,
  \end{EQA}
  where \(\mathbf{Z} \in \RR^{d \t \nsize}\) is an embedding matrix and \(\ev_i\) is a vector with \(1\) on the \(i\)-th place and zeros elsewhere.

  The other possible approach is to treat the local neighborhood of the node as a feature vector and consider
  \begin{EQA}[c]
  \label{general_encoder}
    \encoder_i = f(\av_i),
  \end{EQA}
  where \(\av_i\) is the \(i\)-th column of the adjacency matrix \(\adjacencyMatrix\) and \(f\colon \RR^{\nsize} \to \RR^{d}\) is some function. The particular choice of function \(f\) might be a neural network~\cite{Wang2016,Cao2016}. 

  Also one can consider linear function \(f\):
  \begin{EQA}[c]
  \label{linear_encoder}
    \encoder_i = \embeddingMatrix \av_i + \bv,
  \end{EQA}
  where \(\embeddingMatrix \in \RR^{d \t \nsize}\) is a parameter matrix, \(\bv \in \RR^{d}\) is an intercept vector and a full set of encoder parameters is \(\encoderParams = (\embeddingMatrix, \bv)\). If adjacency matrix \(\adjacencyMatrix\) has rank at least \(d\) then the expressive ability of such an encoder is exactly equal to the one in a direct embedding approach~\eqref{direct_embedding}.

\section{Discrimination of Similarity Distributions}
\label{sec:discriminative_loss}
  In this paper we introduce a different approach which considers a whole set of similarities between nodes and works out a discriminative loss between the distributions of \(\decoder(\encoder_i, \encoder_j)\) in the pairs of similar and non similar nodes. Currently, this approach assumes that the graph is sparse so that similarity matrix \(\similarityMatrix = \bigl[\similarity_{ij}\bigr]_{i, j = 1}^{\nsize}\) is also sparse (we consider either adjacency matrix \(\similarityMatrix = \adjacencyMatrix\) or second order proximities \(\similarityMatrix = \adjacencyMatrix^2\)). Consider the set of all positive similarities
  \(
    L^+ = \{\decoder(\encoder_i, \encoder_j)\colon \similarity_{ij} > 0\}
  \) 
  and the set of all pairs of nodes with zero similarity
  \(
    L^- = \{\decoder(\encoder_i, \encoder_j)\colon \similarity_{ij} = 0\}\).
  
  Our main assumption is that the embedding should allow us to distinguish between similar and non similar nodes. In particular, decoded similarities should be higher for similar nodes. If we treat respective positive similarity \(\similarity_{ij}\) as a weight for the considered decoded similarity value \(\decoder(\encoder_i, \encoder_j)\), then we consider distributions \(\histPos = \histPos(\encoderParams)\) and \(\histNeg = \histNeg(\encoderParams)\) of decoded similarities in \(L^+\) and \(L^-\), respectively, and define the loss function as
  \begin{EQA}[c]
  \label{discriminative_loss}
    \LC(\encoderParams) = -D\bigl(\histPos(\encoderParams), \histNeg(\encoderParams)\bigr),
  \end{EQA}
  where \(D\) is a distance between distributions. It might be KL-divergence, Hellinger distance or, for example, Wasserstein distance. Thus, our aim is to maximize the distance between distributions of positive and negative pairs. 

  We are going to proceed with linear encoder~\eqref{linear_encoder} and suggest to use Pearson correlation as a decoder function \(\decoder\):
  \begin{EQA}[c]
  \label{corr}
    \decoder(\EC_i, \EC_j) =  \frac{\EC_i^{\T} \EC_j}{\|\EC_i\| \|\EC_j\|}.
  \end{EQA}
  In this case \(\decoder(\EC_i, \EC_j) \in [-1, 1]\) which is convenient for our purposes.
    
  For the implementation of discriminative loss~\eqref{discriminative_loss} we follow~\cite{hist_loss} and approximate distributions of decoded similarities in \(L^+\) and \(L^-\) by histogram estimators \(P^+\) and \(P^-\) with linear slope in each bin.
  
  As a distance between distributions we suggest to use 1-D Wasserstein distance (also known as an \textit{``earth mover distance''}, \textit{EMD}) which for the histogram case can be computed as~\cite{Martinez2016}:
  \begin{EQA}[c]
  \label{EMD}
    EMD(\histPos, \histNeg) = \sum_{i = 1}^{N_b} |\varphi_i|,
  \end{EQA}
  where \(N_b\) is the number of bins in the histogram and
  \begin{EQA}[c]
    \varphi_i = \sum_{j = 1}^i \left(\frac{\histPos_{j}}{\lVert \histPos \rVert_1} - \frac{\histNeg_{j}}{\lVert \histNeg \rVert_1}\right).
  \end{EQA}
  However, in our case it is required that the distribution of similar pairs \(\histPos\) should be to the right from the distribution of dissimilar pairs \(\histNeg\). The following simple modification allows to avoid unnecessary local optima:
  \begin{EQA}[c]
  \label{EMD_asym}
    EMD_{asym}(\histPos, \histNeg) = \sum_{i=1}^{N_b} \varphi_i.
  \end{EQA}
  Maximization of modified Wasserstein distance~\eqref{EMD_asym} makes distribution \(\histPos\) concentrating near \(1\), while distribution \(\histNeg\) concentrates near \(-1\). However, it is natural to assume that nodes, which are far away in the graph, should have the embeddings that are independent from each other rather than opposite. That is why we propose to keep in the computation of Wasserstein distance only the part \(\histNeg_{cut}\) of negative distribution \(\histNeg\) corresponding to nonnegative decoded similarities \(\decoder(\encoder_i, \encoder_j) >= 0\). 
    
  Finally, our objective function becomes
  \begin{EQA}[c]
  \label{final_problem}
    \Loss(\encoderParams) = EMD_{asym}(\histNeg_{cut}(\encoderParams), \histPos(\encoderParams)),
  \end{EQA}
  which can be optimized via gradient-based methods to find optimal embedding configuration. The gradient of Wasserstein distance between histograms is easily computable, see~\cite{Martinez2016,hist_loss}.

\section{Experiments}
  In this section we will discuss the experimental evaluation of the proposed algorithm. The whole algorithm was implemented in Tensorflow\footnote{The code of the algorithm and all the experiments is available at~\url{https://github.com/premolab/GraphEmbeddings}.}, while optimization of the functional~\eqref{final_problem} over parameters (embedding matrix \(\embeddingMatrix\) and vector \(\mathbf{b}\)) was performed via stochastic gradient descent.
    
  The further speedup of the algorithm was achieved by subsampling set \(L^-\) (the so-called negative sampling~\cite{Mikolov2013}) which is necessary as real-world graphs are usually sparse with \(|L^-| \gg |L^+|\).

\subsection{Experimental setup}
\label{sec:data}
  For the experiments we used several real-world networks with number of nodes varying from 100 to 4000. The information about the datasets is summarized in the Table~\ref{table_graphs}.
    
  \begin{table}
    \begin{center}
      \begin{tabular}{ccc}
        Name & Number nodes & Number edges  \\
        \noalign{\smallskip}
        \hline
        \noalign{\smallskip}
        Books about US Politics~\cite{Krebs2004} & 105 & 441 \\
        American College Football~\cite{Newman2004} & 115 & 613\\
        Email EU~\cite{Yin2017} & 986 & 25552 \\
        Facebook~\cite{Leskovec2012} & 4039 & 88234
      \end{tabular}
    \end{center}
    \caption{List of real-world networks use in our experiments.}
  \label{table_graphs}
  \end{table}
    
  In our experiments we focus on the \textit{link prediction} problem which offers an universal way to estimate the quality of the embedding for any network as it does not require any additional data except for the graph itself. The solution pipeline starts with constructing the embedding based on the part of graph edges and then checks how well the missing edges can be predicted basing on the embeddings. More precisely, the pipeline is as follows:
    
  \begin{enumerate}
    \item The set of all edges \(\edgeSet\) is randomly divided into two parts: \(\edgeSet_{train}\) and \(\edgeSet_{test}\)\footnote{In our experiments we set \(|\edgeSet_{train}| = |\edgeSet_{test}| = \frac{|\edgeSet|}{2}\).}.
  
    \item The embedding \(\{\EC_v, ~ v \in \nodeSet\}\) is constructed basing on the subgraph \(\GC_{train} = (\nodeSet, \edgeSet_{train})\).
    
    \item Define variables
      \begin{EQA}
        y_{uv}
        &=&
        \begin{cases}
          1, \text{if } (u, v) \in \edgeSet,  \\
          0, \text{if } (u, v) \notin \edgeSet,  \\
        \end{cases}
        \\
        X_{uv} &=& \langle\EC_u, \EC_v\rangle,
      \end{EQA}
      where \(\langle\cdot\,, \cdot\rangle\) means concatenation. 
        
    \item Construct classifier (logistic regression in our experiments) basing on the training data set \(\bigl\{(X_{uv}, y_{uv}), (u, v) \in \edgeSet_{train}\}\) and estimate the classification quality on the test set \(\bigl\{(X_{uv}, y_{uv}), (u, v) \in \edgeSet_{test}\bigr\}\).
  \end{enumerate}
    
\subsection{Results}
\label{sec:results}
  We compare the results of the proposed algorithm (DDoS) with several state-of-the-art algorithms belonging to different subcategories of embedding algorithms: random walk based algorithms (DeepWalk~\cite{deepwalk}), direct matrix factorization approaches (HOPE~\cite{HOPE}) and a neural network based autoencoders (SDNE~\cite{sdne}).

\begin{figure}[h!]
    \centering
    \includegraphics[width=.95\linewidth]{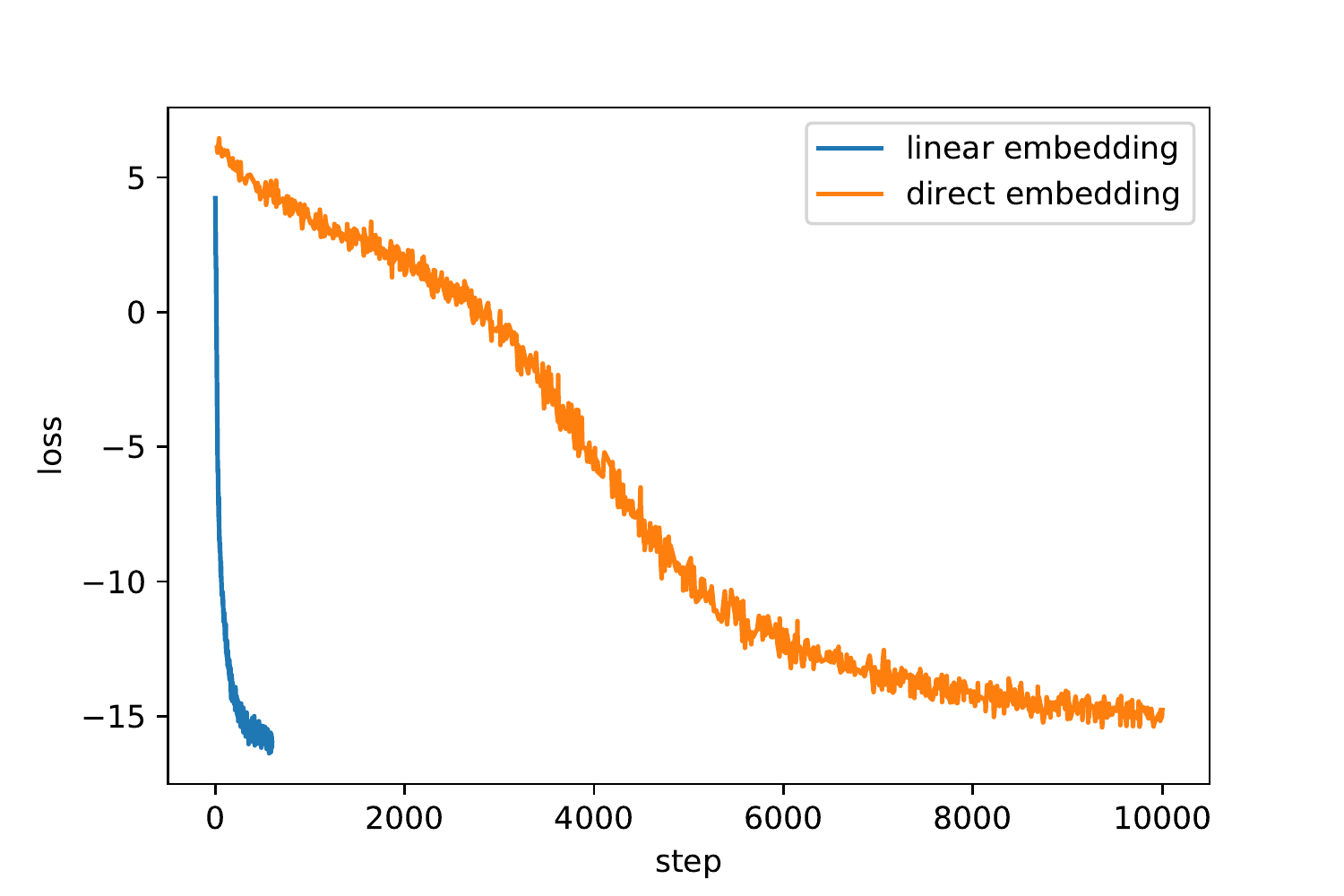}
    \caption{Steps of loss optimization for the parametrized encoder~\eqref{linear_encoder} and the embedding lookup~\eqref{direct_embedding} in orange and blue respectively.}
  \label{convergence}
  \end{figure}

  The results are summarized in Table~\ref{tab:results}. As we  can see, in the majority of cases DDoS embeddings allow us to achieve better results than their competitors. Interestingly the usage of the parametrized encoder~\eqref{linear_encoder} instead of the embedding lookup~\eqref{direct_embedding} has resulted in a much faster convergence of the algorithm, see Figure~\ref{convergence}.

  \begin{table}
    \begin{tabular}{|c|c|c|c|c|c|}
      \cline{1-6}
      Dataset & \(d\) & HOPE & DeepWalk & SDNE & DDoS \\ \cline{1-6}

      Books& 4  & 0.89 & 0.87 & 0.78 & \textbf{0.90} \\ \cline{2-6}
      about & 8  & \textbf{0.90} & 0.89 & 0.79 & \textbf{0.90} \\ \cline{2-6}
      US Politics & 16 & \textbf{0.92} & 0.90 & 0.81 & 0.90 \\ \cline{2-6}
      & 32 & \textbf{0.93} & 0.90 & 0.75 & 0.90 \\ \cline{1-6}

      American & 4  & 0.75 & 0.85 & 0.77 & \textbf{0.87} \\ \cline{2-6}
      College & 8  & 0.86 & \textbf{0.90} & 0.82 & \textbf{0.90} \\ \cline{2-6}
      Football & 16 & 0.90 & \textbf{0.91} & 0.84 & \textbf{0.91} \\ \cline{2-6}
      & 32 & 0.92 & 0.92 & 0.84 & \textbf{0.93} \\ \cline{1-6}

      & 4  & 0.74 & 0.81 & 0.89 & \textbf{0.90} \\ \cline{2-6}
      Email EU & 8  & 0.83 & 0.86 & 0.90 & \textbf{0.92} \\ \cline{2-6}
      & 16 & 0.90 & 0.89 & 0.92 & \textbf{0.93} \\ \cline{2-6}
      & 32 & \textbf{0.93} & 0.90 & \textbf{0.93} & \textbf{0.93} \\ \cline{1-6}

      & 4  & 0.84 & 0.94 & 0.96 & \textbf{0.97} \\ \cline{2-6}
      Facebook & 8  & 0.92 & \textbf{0.98} & 0.96 & \textbf{0.98} \\ \cline{2-6}
      & 16 & 0.96 & \textbf{0.99} & 0.97 & \textbf{0.99} \\ \cline{2-6}
      & 32 & 0.97 & 0.99 & 0.98 & \textbf{1.00} \\ \cline{1-6}    
    \end{tabular}
    \vspace{5pt}
    \caption{Link prediction results (ROC-AUC values).}
  \label{tab:results}
  \end{table}

\section{Conclusions and Outlook}
\label{sec:conclusions}
  In this work we propose a simple but powerful approach for constructing the graph embeddings based on discrimination of similarity distributions. We show the way to implement the general idea by using maximization of a specially tuned Wasserstein distance. The series of experiments with the link prediction in real-world graphs convincingly demonstrate the superiority of the proposed approach over its competitors. 

  Our works offers a number of directions for further development. Among them is the scalability of the algorithm that should be improved first of all. It can be achieved by combining the proposed criterion with sampling graph nodes via random walks. The other direction is to test DDoS embeddings in other network analysis tasks such as community detection and semi-supervised node classification. Extensions to directed and weighted graphs also seem to be of a great interest. Finally, the usage of a non-linear encoder~\eqref{general_encoder} parametrized by a neural network can be a promising direction for further investigation and improvement.

\bibliographystyle{plain}
\bibliography{embeddings}

\end{document}